\documentclass{article}
\pdfoutput=1

\usepackage{graphicx}
\usepackage{subfigure}

\usepackage[accepted]{icml2018}

\PassOptionsToPackage{numbers, compress}{natbib}

\usepackage[utf8]{inputenc} 
\usepackage[T1]{fontenc}    

\usepackage{xcolor}
\definecolor{nice-red}{HTML}{E41A1C}
\colorlet{dark-red}{nice-red!80!black}
\definecolor{nice-orange}{HTML}{FF7F00}
\colorlet{dark-orange}{orange!85!black}
\definecolor{nice-yellow}{HTML}{FFC020}
\definecolor{nice-green}{HTML}{4DAF4A}
\definecolor{nice-blue}{HTML}{377EB8}
\definecolor{nice-purple}{HTML}{984EA3}

\usepackage{hyperref}       
\hypersetup{
 unicode=false,          
 pdftoolbar=true,        
 pdfmenubar=true,        
 pdffitwindow=false,     
 pdfstartview={FitH},    
 pdftitle={Towards Neural Theorem Proving at Scale},    
 pdfauthor={Pasquale Minervini},     
 pdfnewwindow=true,      
 colorlinks=true,        
 linkcolor=black,        
 citecolor=blue!50!black,    
 filecolor=black,        
 urlcolor=black          
}
\usepackage{url}            
\usepackage{booktabs}       
\usepackage{nicefrac}       
\usepackage{microtype}      

\usepackage[xindy]{glossaries}

\usepackage[]{todonotes}

\usepackage{bm}
\usepackage{cleveref} 
\crefformat{equation}{Eq.~#2#1#3}
\Crefformat{equation}{Equation~#2#1#3}
\crefrangeformat{equation}{Eqs.~#3#1#4 to~#5#2#6}
\Crefrangeformat{equation}{Equations~#3#1#4 to~#5#2#6}
\crefmultiformat{equation}{Eqs.~#2#1#3}%
{ and~#2#1#3}{, #2#1#3}{ and~#2#1#3}
\Crefmultiformat{equation}{Equations~#2#1#3}%
{ and~#2#1#3}{, #2#1#3}{ and~#2#1#3}

\usepackage{tikz}
\usetikzlibrary{calc,trees,positioning,arrows,chains,shapes.geometric,%
  decorations.pathreplacing,decorations.pathmorphing,shapes,%
  matrix,shapes.symbols,fit,decorations,arrows.meta}
 
\usepackage{wrapfig}

\usepackage{comment}

\usepackage{amsmath}
\usepackage{amsthm}

\theoremstyle{definition}
\newtheorem{example}{Example}[section]

\usepackage{lipsum}
\usepackage{multirow}

\usepackage{amssymb}
\usepackage{soul}

\newacronym{AKBC}{AKBC}{Automated Knowledge Base Construction}
\newacronym{AUC}{AUC}{Area Under Curve}

\newacronym{BPR}{BPR}{Bayesian Personalized Ranking}

\newacronym{CNF}{CNF}{Conjunctive Normal Form}
\newacronym{CNN}{CNN}{Convolutional Neural Network}
\newacronym{CWA}{CWA}{Closed World Assumption}

\newacronym{FOIL}{FOIL}{First Order Inductive Learner}

\newacronym{GPCA}{GPCA}{Generalized Principal Component Analysis}
\newacronym{GPU}{GPU}{Graphics Processing Unit}

\newacronym{ILP}{ILP}{Inductive Logic Programming}

\newacronym{KB}{KB}{Knowledge Base}

\newacronym[longplural={Long Short-Term Memories}]{LSTM}{LSTM}{long short-term memory}

\newacronym{NLP}{NLP}{Natural Language Processing}
\newacronym{NLI}{NLI}{Natural Language Inference}

\newacronym{NTP}{NTP}{Neural Theorem Prover}

\newacronym{NTN}{NTN}{Neural Tensor Network}

\newacronym{MAP}{MAP}{Mean Average Precision}
\newacronym{MLP}{MLP}{Multi-layer Perceptron}
\newacronym{MRR}{MRR}{Mean Reciprocal Rank}

\newacronym{OpenIE}{OpenIE}{Open Information Extraction}

\newacronym{PCA}{PCA}{Principal Component Analysis}
\newacronym{PRA}{PRA}{Path Ranking Algorithm}
\newacronym{ProPPR}{ProPPR}{Programming with Personalized PageRank}

\newacronym{RBF}{RBF}{Radial Basis Function}
\newacronym{ROC}{ROC}{Receiver Operating Characteristic}
\newacronym{RNN}{RNN}{Recurrent Neural Network}
\newacronym{RTE}{RTE}{Recognizing Textual Entailment}

\newacronym{SGD}{SGD}{Stochastic Gradient Descent}
\newacronym{SNLI}{SNLI}{Stanford Natural Language Inference}

\newacronym{ANNS}{ANNS}{Approximate Nearest Neighbour Search}

\newacronym{LSH}{LSH}{Locality-Sensitive Hashing}
\newacronym{PQ}{PQ}{Product Quantization}
\newacronym{PG}{PG}{Proximity Graph}
\newacronym{HNSW}{HNSW}{Hierarchical Navigable Small World}

\newcommand{\kb}{\mathfrak{K}}
\newcommand{\state}{S}
\renewcommand{\emptyset}{\varnothing}
\newcommand{\fail}{\verb~FAIL~}
\newcommand{\fun}[1]{\text{#1}}

\let\union\cup

\let\set\dom 
\newcommand{\lss}[1]{\mathbb{#1}}
\let\ls\textsc
\let\lst\ls
\newcommand{\module}[1]{{\color{nice-purple!75!black}\verb~#1~}}

\newcommand{\xs}[1]{\bm{[} #1 \bm{]}}
\newcommand{\emptylist}{\bm{[}\ \bm{]}}

\renewcommand\vec[1]{{\bm{#1}}}

\makeatletter
\newcommand*\bdot{\mathpalette\bdot@{.5}}
\newcommand*\bdot@[2]{\mathbin{\vcenter{\hbox{\scalebox{#2}{$\m@th#1\bullet$}}}}}
\makeatother

\def\params{{\vec{\theta}}}

\DeclareMathOperator*{\argmax}{arg\,max}

\newcommand{\rel}[1]{\verb~#1~}
\newcommand{\const}[1]{\textsc{#1}}
\let\ent\const
\newcommand{\var}[1]{\textsc{#1}}
\def\lif{\ \text{:--}\ }

\let\subs\psi

\newcommand{\pred}[1]{\text{$\verb~#1~$}}

\let\success\rho

\let\todonote\todo

\colorlet{fixme}{red!85!black}
\colorlet{fixme-bright}{fixme!25}

\colorlet{todo}{orange!85!black}
\colorlet{todo-bright}{todo!25}

\colorlet{review}{nice-yellow!50!nice-orange}
\colorlet{review-bright}{review!25}

\colorlet{maybe}{nice-yellow}
\colorlet{maybe-bright}{maybe!25}

\colorlet{toref}{purple!70!black}
\colorlet{toref-bright}{toref!25}

\colorlet{info}{green!50!black}
\colorlet{info-bright}{info!25}

\colorlet{cut}{black!60}
\colorlet{cut-bright}{cut!25}

\colorlet{extend}{blue!85!black}
\colorlet{extend-bright}{extend!25}

\colorlet{discuss}{nice-red!50!black}
\colorlet{discuss-bright}{discuss!25}

\renewcommand{\todo}[1]{\todonote[linecolor=todo, backgroundcolor=todo-bright, bordercolor=todo]{\thesection{} {\bf\color{todo}TODO} #1}{}}

\newcommand{\info}[1]{}
\newcommand{\hlinfo}[2]{}

\newcommand{\maybe}[1]{}
\newcommand{\hlmaybe}[2]{}

\newcommand{\at}[3]{
  \begin{scope}[shift={(#1,#2)}]
    #3
  \end{scope} 
}

\newcommand{\scale}[2]{
  \begin{scope}[scale=#1]
    #2
  \end{scope}
}

\newcommand{\name}[2]{
  \begin{scope}[local bounding box=#1]
    #2
  \end{scope}
}

\newcommand{\srep}[2]{
  \pgfmathsetseed{#2}
  \begin{scope}[scale=0.5]
    \begin{scope}[shift={(-#1/2+0.5,-1/2.0)}]
      \draw[ultra thick, fill=white] (-0.5, 0) rectangle (#1-0.5, 1);
      \foreach \y in {1,...,1} {
        \foreach \x in {1,...,#1} {
          \pgfmathsetmacro\c{(rand+1)*50}
          \draw[very thick, fill=nice-blue!\c!nice-red] (\x-1, \y-0.5) circle (0.44);
        }
      }
    \end{scope}
  \end{scope}
}

\newcommand{\stateone}{
  \scale{0.75}{
    \draw[ultra thick] (0,0) rectangle (4,1.5);
    \draw[ultra thick, dashed] (2,0) -- (2,1.5);
    \node[text width=2cm, text centered] at (1,0.75) {
      \huge
      $\emptyset$
    };

    \at{0}{-1}{
      \name{c1}{\draw[thick, fill=white] (3,2.25) circle (0.1);}
      \name{c2}{\draw[thick, fill=white] (2.5,2) circle (0.1);}
      
      \name{c3}{\draw[thick, fill=white] (3,1.75) circle (0.1);}
      \name{c4}{\draw[thick, fill=white] (3.5,1.5) circle (0.1);}
      \name{c5}{\draw[thick, fill=white] (3,1.25) circle (0.1);}

      \draw[thick, -Latex] (c1) -- (c3);
      \draw[thick, -Latex] (c2) -- (c3);
      \draw[thick, -Latex] (c3) -- (c5);
      \draw[thick, -Latex] (c4) -- (c5);
    }
  }
}

\newcommand{\statethree}{
  \scale{0.75}{
    \draw[ultra thick] (0,0) rectangle (4,1.5);
    \draw[ultra thick, dashed] (2,0) -- (2,1.5);
    \node[text width=2cm, text centered] at (1,0.75) {
      \scriptsize
      \var{X}/\const{abe}\\[-0.4em]
      \var{Y}/\const{bart}
    };

    \at{0}{-1}{
      \name{c1}{\draw[thick, fill=white] (3,2.25) circle (0.1);}
      \name{c2}{\draw[thick, fill=white] (2.5,2) circle (0.1);}
      
      \name{c3}{\draw[thick, fill=white] (3,1.75) circle (0.1);}
      \name{c4}{\draw[thick, fill=white] (3.5,1.5) circle (0.1);}
      \name{c5}{\draw[thick, fill=white] (3,1.25) circle (0.1);}

      \draw[thick, -Latex] (c1) -- (c3);
      \draw[thick, -Latex] (c2) -- (c3);
      \draw[thick, -Latex] (c3) -- (c5);
      \draw[thick, -Latex] (c4) -- (c5);
    }
  }
}

\newcommand{\statethreeone}{
  \scale{0.75}{
    \draw[ultra thick] (0,0) rectangle (4,2.5);
    \draw[ultra thick, dashed] (2,0) -- (2,2.5);
    \node[text width=2cm, text centered] at (1,1.25) {
      \scriptsize
      \var{X}/\const{abe}\\
      \var{Y}/\const{bart}\\[-0.4em]
      \var{Z}/\const{homer}
    };

    \name{c1}{\draw[thick, fill=white] (3,2.25) circle (0.1);}
    \name{c2}{\draw[thick, fill=white] (2.5,2) circle (0.1);}
    
    \name{c3}{\draw[thick, fill=white] (3,1.75) circle (0.1);}
    \name{c4}{\draw[thick, fill=white] (3.5,1.5) circle (0.1);}
    \name{c5}{\draw[thick, fill=white] (3,1.25) circle (0.1);}

    \name{c6}{\draw[thick, fill=white] (2.5,1) circle (0.1);}
    \name{c7}{\draw[thick, fill=white] (3,.75) circle (0.1);}
    \name{c8}{\draw[thick, fill=white] (3.5,.5) circle (0.1);}
    \name{c9}{\draw[thick, fill=white] (3,.25) circle (0.1);}

    \draw[thick, -Latex] (c1) -- (c3);
    \draw[thick, -Latex] (c2) -- (c3);
    \draw[thick, -Latex] (c3) -- (c5);
    \draw[thick, -Latex] (c4) -- (c5);
    \draw[thick, -Latex] (c5) -- (c7);
    \draw[thick, -Latex] (c6) -- (c7);
    \draw[thick, -Latex] (c7) -- (c9);
    \draw[thick, -Latex] (c8) -- (c9);
  }
}

\newcommand{\statethreetwo}{
  \scale{0.75}{
    \draw[ultra thick] (0,0) rectangle (4,2.5);
    \draw[ultra thick, dashed] (2,0) -- (2,2.5);
    \node[text width=2cm, text centered] at (1,1.25) {
      \scriptsize
      \var{X}/\const{abe}\\
      \var{Y}/\const{bart}\\[-0.4em]
      \var{Z}/\const{bart}
    };

    \name{c1}{\draw[thick, fill=white] (3,2.25) circle (0.1);}
    \name{c2}{\draw[thick, fill=white] (2.5,2) circle (0.1);}
    
    \name{c3}{\draw[thick, fill=white] (3,1.75) circle (0.1);}
    \name{c4}{\draw[thick, fill=white] (3.5,1.5) circle (0.1);}
    \name{c5}{\draw[thick, fill=white] (3,1.25) circle (0.1);}

    \name{c6}{\draw[thick, fill=white] (2.5,1) circle (0.1);}
    \name{c7}{\draw[thick, fill=white] (3,.75) circle (0.1);}
    \name{c8}{\draw[thick, fill=white] (3.5,.5) circle (0.1);}
    \name{c9}{\draw[thick, fill=white] (3,.25) circle (0.1);}

    \draw[thick, -Latex] (c1) -- (c3);
    \draw[thick, -Latex] (c2) -- (c3);
    \draw[thick, -Latex] (c3) -- (c5);
    \draw[thick, -Latex] (c4) -- (c5);
    \draw[thick, -Latex] (c5) -- (c7);
    \draw[thick, -Latex] (c6) -- (c7);
    \draw[thick, -Latex] (c7) -- (c9);
    \draw[thick, -Latex] (c8) -- (c9);
  }
}

\newcommand{\statethreeoneone}{
  \scale{0.75}{
    \draw[ultra thick] (0,-1) rectangle (4,2.5);
    \draw[ultra thick, dashed] (2,-1) -- (2,2.5);
    \node[text width=2cm, text centered] at (1,0.75) {
      \scriptsize
      \var{X}/\const{abe}\\
      \var{Y}/\const{bart}\\[-0.4em]
      \var{Z}/\const{homer}
    };

    \name{c1}{\draw[thick, fill=white] (3,2.25) circle (0.1);}
    \name{c2}{\draw[thick, fill=white] (2.5,2) circle (0.1);}
    
    \name{c3}{\draw[thick, fill=white] (3,1.75) circle (0.1);}
    \name{c4}{\draw[thick, fill=white] (3.5,1.5) circle (0.1);}
    \name{c5}{\draw[thick, fill=white] (3,1.25) circle (0.1);}

    \name{c6}{\draw[thick, fill=white] (2.5,1) circle (0.1);}
    \name{c7}{\draw[thick, fill=white] (3,.75) circle (0.1);}
    \name{c8}{\draw[thick, fill=white] (3.5,.5) circle (0.1);}
    \name{c9}{\draw[thick, fill=white] (3,.25) circle (0.1);}

    \name{c10}{\draw[thick, fill=white] (2.5,0) circle (0.1);}
    \name{c11}{\draw[thick, fill=white] (3,-.25) circle (0.1);}
    \name{c12}{\draw[thick, fill=white] (3.5,-.55) circle (0.1);}
    \name{c13}{\draw[thick, fill=white] (3,-.75) circle (0.1);}

    \draw[thick, -Latex] (c1) -- (c3);
    \draw[thick, -Latex] (c2) -- (c3);
    \draw[thick, -Latex] (c3) -- (c5);
    \draw[thick, -Latex] (c4) -- (c5);
    \draw[thick, -Latex] (c5) -- (c7);
    \draw[thick, -Latex] (c6) -- (c7);
    \draw[thick, -Latex] (c7) -- (c9);
    \draw[thick, -Latex] (c8) -- (c9);
    \draw[thick, -Latex] (c9) -- (c11);
    \draw[thick, -Latex] (c10) -- (c11);
    \draw[thick, -Latex] (c11) -- (c13);
    \draw[thick, -Latex] (c12) -- (c13);          
  }
}

\newcommand{\statethreetwoone}{
  \scale{0.75}{
    \draw[ultra thick] (0,-1) rectangle (4,2.5);
    \draw[ultra thick, dashed] (2,-1) -- (2,2.5);
    \node[text width=2cm, text centered] at (1,0.75) {
      \scriptsize
      \var{X}/\const{abe}\\
      \var{Y}/\const{bart}\\[-0.4em]
      \var{Z}/\const{bart}
    };

    \name{c1}{\draw[thick, fill=white] (3,2.25) circle (0.1);}
    \name{c2}{\draw[thick, fill=white] (2.5,2) circle (0.1);}
    
    \name{c3}{\draw[thick, fill=white] (3,1.75) circle (0.1);}
    \name{c4}{\draw[thick, fill=white] (3.5,1.5) circle (0.1);}
    \name{c5}{\draw[thick, fill=white] (3,1.25) circle (0.1);}

    \name{c6}{\draw[thick, fill=white] (2.5,1) circle (0.1);}
    \name{c7}{\draw[thick, fill=white] (3,.75) circle (0.1);}
    \name{c8}{\draw[thick, fill=white] (3.5,.5) circle (0.1);}
    \name{c9}{\draw[thick, fill=white] (3,.25) circle (0.1);}

    \name{c10}{\draw[thick, fill=white] (2.5,0) circle (0.1);}
    \name{c11}{\draw[thick, fill=white] (3,-.25) circle (0.1);}
    \name{c12}{\draw[thick, fill=white] (3.5,-.55) circle (0.1);}
    \name{c13}{\draw[thick, fill=white] (3,-.75) circle (0.1);}

    \draw[thick, -Latex] (c1) -- (c3);
    \draw[thick, -Latex] (c2) -- (c3);
    \draw[thick, -Latex] (c3) -- (c5);
    \draw[thick, -Latex] (c4) -- (c5);
    \draw[thick, -Latex] (c5) -- (c7);
    \draw[thick, -Latex] (c6) -- (c7);
    \draw[thick, -Latex] (c7) -- (c9);
    \draw[thick, -Latex] (c8) -- (c9);
    \draw[thick, -Latex] (c9) -- (c11);
    \draw[thick, -Latex] (c10) -- (c11);
    \draw[thick, -Latex] (c11) -- (c13);
    \draw[thick, -Latex] (c12) -- (c13);          
  }
}

\renewcommand{\rel}[1]{{\color{nice-purple!75!black}\verb~#1~}}
\renewcommand{\const}[1]{{\color{nice-blue!75!black}\textsc{#1}}}
\let\ent\const
\renewcommand{\var}[1]{{\color{nice-orange!75!black}\textsc{#1}}}
\def\lif{\ {\color{nice-red!75!black}\text{:--}}\ }

\renewcommand{\var}[1]{{\color{nice-orange!75!black}\textsc{#1}}}
\def\lif{\ {\color{nice-red!75!black}\text{:--}}\ }

\newcommand{\node}{\ensuremath{n}}

\DeclareMathOperator{\krnl}{k}

\renewcommand{\kb}{\ensuremath{\mathcal{K}}}

\icmltitlerunning{Towards Neural Theorem Proving at Scale}

\begin{document}

\twocolumn[
\icmltitle{Towards Neural Theorem Proving at Scale}

\icmlsetsymbol{equal}{*}

\begin{icmlauthorlist}
\icmlauthor{Pasquale Minervini*}{UCL}
\icmlauthor{Matko Bosnjak*}{UCL}
\icmlauthor{Tim Rockt{\"{a}}schel}{Oxford}
\icmlauthor{Sebastian Riedel}{UCL,BAI}
\end{icmlauthorlist}

\icmlaffiliation{UCL}{Department of Computer Science, University College London, London, United Kingdom}
\icmlaffiliation{Oxford}{University of Oxford, Oxford, United Kingdom}
\icmlaffiliation{BAI}{Bloomsbury AI, United Kingdom}

\icmlcorrespondingauthor{Pasquale Minervini}{p.minervini@cs.ucl.ac.uk}

\icmlkeywords{Neural Theorem Provers, Approximate Nearest Neighbours, Neural-Symbolic Integration, Program Induction}

\vskip 0.3in
]

\printAffiliationsAndNotice{\icmlEqualContribution}

\begin{abstract}
Neural models combining representation learning and reasoning in an end-to-end trainable manner are receiving increasing interest.
However, their use is severely limited by their computational complexity, which renders them unusable on real world datasets.
We focus on the \gls{NTP} model proposed by \citet{rocktaschel2017end}, a continuous relaxation of the Prolog backward chaining algorithm where unification between terms is replaced by the similarity between their embedding representations.
For answering a given query, this model needs to consider all possible proof paths, and then aggregate results -- this quickly becomes infeasible even for small \glspl{KB}.
We observe that we can accurately approximate the inference process in this model by considering only proof paths associated with the highest proof scores.
This enables inference and learning on previously impracticable \glspl{KB}.
\end{abstract}
\section{Introduction} \label{sec:introduction}

Recent advancements in deep learning intensified the long-standing interests in integrating symbolic reasoning with connectionist models~\citep{shen1988theoretical,ding1996prolog,garcez2012neural,DBLP:journals/corr/abs-1801-00631}.
The attraction of said integration stems from the complementing properties of these systems.
Symbolic reasoning models offer interpretability, efficient generalisation from a small number of examples, and the ability to leverage knowledge provided by an expert.
However, these systems are unable to handle ambiguous and noisy high-dimensional data such as sensory inputs~\citep{DBLP:conf/ilp/RaedtK08}.
On the other hand, representation learning models exhibit robustness to noise and ambiguity, can learn task-specific representations, and achieve state-of-the-art results on a wide variety of tasks~\citep{DBLP:journals/pami/BengioCV13}.
However, being universal function approximators, these models require vast amounts of training data and are treated as non-interpretable \emph{black boxes}.
One way of integrating the symbolic and sub-symbolic models is by continuously relaxing discrete operations and implementing them in a connectionist framework.
Recent approaches in this direction focused on learning algorithmic behaviour without the explicit symbolic representations of a program~\citep{DBLP:journals/corr/GravesWD14,DBLP:journals/nature/GravesWRHDGCGRA16,DBLP:journals/corr/KaiserS15,DBLP:journals/corr/NeelakantanLS15,DBLP:journals/corr/AndrychowiczK16}, and consequently with it~\citep{DBLP:journals/corr/ReedF15,DBLP:conf/icml/BosnjakRNR17,gaunt2016terpret,parisotto2016neuro}.
In the inductive logic programming setting, two new models, \glspl{NTP}~\citep{rocktaschel2017end} and Differentiable Inductive Logic Programming ({\texorpdfstring{$\partial$ILP}{dILP}})~\citep{evans2018learning} successfully combined the interpretability and data efficiency of a logic programming system with the expressiveness and robustness of neural networks.
In this paper, we focus on the \gls{NTP} model proposed by \citet{rocktaschel2017end}.
Akin to recent neural-symbolic models, \glspl{NTP} rely on a continuous relaxation of a discrete algorithm, operating over the sub-symbolic representations.
In this case, the algorithm is an analogue to Prolog's backward chaining with a relaxed unification operator.
The backward chaining algorithm constructs neural networks, which model continuously relaxed proof paths using sub-symbolic representations.
These representations are learned end-to-end by maximising the proof scores of facts in the \gls{KB}, while minimising the score of facts not in the \gls{KB}, in a link prediction setting~\citep{DBLP:journals/pieee/Nickel0TG16}.
However, while the symbolic unification checks whether two terms can represent the same structure, the relaxed unification measures the similarity between their sub-symbolic representations.
\begin{figure*}[ht!]
  \centering    
  \resizebox{\textwidth}{!}{
    \begin{tikzpicture}
      \node[anchor = north west, text width=7cm] at (-12, 1.25) {
        Example Knowledge Base:\\
        \fcolorbox{black}{nice-green!50}{1.} $\rel{fatherOf}(\ent{abe}, \ent{homer}).$\\
        \fcolorbox{black}{nice-yellow!50}{2.} $\rel{parentOf}(\ent{homer}, \ent{bart}).$\\
        \fcolorbox{black}{nice-red!50}{3.} $\rel{grandfatherOf}(\var{X}, \var{Y}) \lif$\\
        $\qquad\quad\rel{fatherOf}(\var{X}, \var{Z}),$\\
        $\qquad\quad\rel{parentOf}(\var{Z}, \var{Y}).$
      };

      \at{-1.5}{0.5}{
        \scale{0.75}{
          \draw[ultra thick] (0,0) rectangle (4,1);
          \draw[ultra thick, dashed] (2,0) -- (2,1);
          \node[text width=2cm, text centered] at (1,0.5) {
            \huge
            $\emptyset$
          };
        }
        \node[] at (2.25,0.4) {\Large $1.0$};          
      }
      
      \name{q}{
        \at{0}{-0.5}{
          \draw[ultra thick, rounded corners, dotted] (-2.8, -0.3) rectangle (2.8, 0.3);
        }
        \name{qp}{\at{-2}{-0.5}{\srep{3}{1}}}
        \name{qc1}{\at{0}{-0.5}{\srep{3}{2}}}
        \name{qc2}{\at{2}{-0.5}{\srep{3}{3}}}
      }
      
      \node[above = 0.02cm of qp] {\rel{grandpaOf}};
      \node[above = 0.1cm of qc1] {\const{abe}};
      \node[above = 0.1cm of qc2] {\const{bart}};
      
      \name{s1}{\at{-7}{-3}{\stateone}}
      \draw[line width=3pt, -Latex] (qc1) --node[pos=0.5, draw, line width=0.5pt, fill=nice-green!50]{1.} (s1);

      \name{s2}{\at{-1.5}{-3}{\stateone}}
      \draw[line width=3pt,-Latex] (qc1) --node[pos=0.4, draw, line width=0.5pt, fill=nice-yellow!50]{2.} (0,-1.8);

      \name{s3}{\at{4}{-3}{\statethree}}
      \draw[line width=3pt,-Latex] (qc1) --node[pos=0.5, draw, line width=0.5pt, fill=nice-red!50]{3.} (s3);
      \at{7.25}{-2.5}{
        \node[text width=4cm, anchor=west] (subgoal0) at (0,0) {3.1 $\rel{fatherOf}(\var{X}, \var{Z})$\\ 3.2 $\rel{parentOf}(\var{Z}, \var{Y})$};
      }
      
      \name{b1}{    
        \at{0}{-4.5}{
          \draw[ultra thick, rounded corners, dotted] (-2.8, -0.3) rectangle (2.3, 0.3);
          \name{b1p}{\at{-2}{0}{\srep{3}{4}}}
          \node[above = 0.1cm of b1p] {\rel{fatherOf}};
          \name{b1c1}{\at{0}{0}{\srep{3}{2}}}
          \node[above = 0.1cm of b1c1] {\const{abe}};
          \node[] at (1.5, 0) {\var{Z}};   
        }
      }
      \path[line width=3pt, dotted, black!50] (subgoal0) edge[bend left=15] (b1);

      \name{s31}{\at{-7}{-7}{\statethreeone}}
      \at{-7}{-6}{
        \node[anchor=east] (subgoal2) at (0,0) {3.2 $\rel{parentOf}(\var{Z}, \var{Y})$};
      }
      \draw[line width=3pt, -Latex] (b1) --node[pos=0.5, draw, line width=0.5pt, fill=nice-green!50]{1.} (s31);

      \name{s32}{\at{4}{-7}{\statethreetwo}}
      \at{7.25}{-6}{
        \node[anchor=west] (subgoal1) at (0,0) {3.2 $\rel{parentOf}(\var{Z}, \var{Y})$};
      }
      \draw[line width=3pt,-Latex] (b1) --node[pos=0.5, draw, line width=0.5pt, fill=nice-yellow!50]{2.} (s32);
      
      \node[] (s33) at (0, -6.5) {\color{nice-red}\fail};
      \draw[line width=3pt,-Latex] (0,-4.7) --node[pos=0.4, draw, line width=0.5pt, fill=nice-red!50]{3.} (s33);

      \at{0}{-0.5}{
        \name{b2}{
          \at{-5.5}{-8}{
            \draw[ultra thick, rounded corners, dotted] (-2.8, -0.3) rectangle (2.8, 0.3);
            \name{b2p}{\at{-2}{0}{\srep{3}{5}}}
            \node[above = 0.02cm of b2p] {\rel{parentOf}};
            \name{b2c1}{\at{0}{0}{\srep{3}{6}}}
            \node[above = 0.1cm of b2c1] {\const{homer}};
            \name{b2c2}{\at{2}{0}{\srep{3}{3}}}
            \node[above = 0.1cm of b2c2] {\const{bart}};
          }
        }

        \path[line width=3pt, dotted, black!50] (subgoal2) edge[bend right] (b2);
        \name{s311}{\at{-9.5}{-11}{\statethreeoneone}}
        \name{s312}{\at{-4.5}{-11}{\statethreeoneone}}
        \node[] (s313) at (-5.6, -10.5) {\color{nice-red}\fail};
        \draw[line width=3pt, -Latex] (b2) --node[pos=0.5, draw, line width=0.5pt, fill=nice-green!50]{1.} (s311);
        \draw[line width=3pt,-Latex] (b2) --node[pos=0.4,  draw, line width=0.5pt, fill=nice-red!50]{3.} (s313);
        \draw[line width=3pt,-Latex] (b2) --node[pos=0.5,  draw, line width=0.5pt, fill=nice-yellow!50]{2.} (s312);
      }
      
      \at{0}{-0.5}{
        \name{b3}{
          \at{5.5}{-8}{
            \draw[ultra thick, rounded corners, dotted] (-2.8, -0.3) rectangle (2.8, 0.3);
            \name{b2p}{\at{-2}{0}{\srep{3}{5}}}
            \node[above = 0.02cm of b2p] {\rel{parentOf}};
            \name{b2c1}{\at{0}{0}{\srep{3}{3}}}
            \node[above = 0.1cm of b2c1] {\const{bart}};
            \name{b2c2}{\at{2}{0}{\srep{3}{3}}}
            \node[above = 0.1cm of b2c2] {\const{bart}};
          }
        }
        \path[line width=3pt, dotted, black!50] (subgoal1) edge[bend left] (b3); 
        \name{s321}{\at{1.5}{-11}{\statethreetwoone}}
        \name{s322}{\at{6.5}{-11}{\statethreetwoone}}
        \node[] (s323) at (5.5, -10.5) {\color{nice-red}\fail};
        \draw[line width=3pt, -Latex] (b3) --node[pos=0.5,  draw, line width=0.5pt, fill=nice-green!50]{1.} (s321);
        \draw[line width=3pt,-Latex] (b3) --node[pos=0.4,  draw, line width=0.5pt, fill=nice-red!50]{3.} (s323);
        \draw[line width=3pt,-Latex] (b3) --node[pos=0.5,  draw, line width=0.5pt, fill=nice-yellow!50]{2.} (s322);
      }

      \at{-3.5}{-2.25}{
        \draw[line width=3pt, nice-green, opacity=0.8] (-7.8,2.5) -- (-3.5,2.5);
        \draw[line width=3pt, nice-yellow, opacity=0.8] (-7.8,2) -- (-3.4,2);
        \draw[line width=3pt, nice-red, opacity=0.8] (-7.8,1.45) -- (-3.9,1.45);
      } 
    \end{tikzpicture}
  }
  \caption{
  A visual depiction of the \gls{NTP}' recursive computation graph construction, applied to a toy \gls{KB} (top left).
  Dash-separated rectangles denote proof states (left: substitutions, right: proof score -generating neural network).
  All the non-\texttt{FAIL} proof states are aggregated to obtain the final proof success (depicted in \Cref{fig:aggregation}).
  Colours and indices on arrows correspond to the respective \gls{KB} rule application.}
  \label{fig:overview}
\end{figure*}
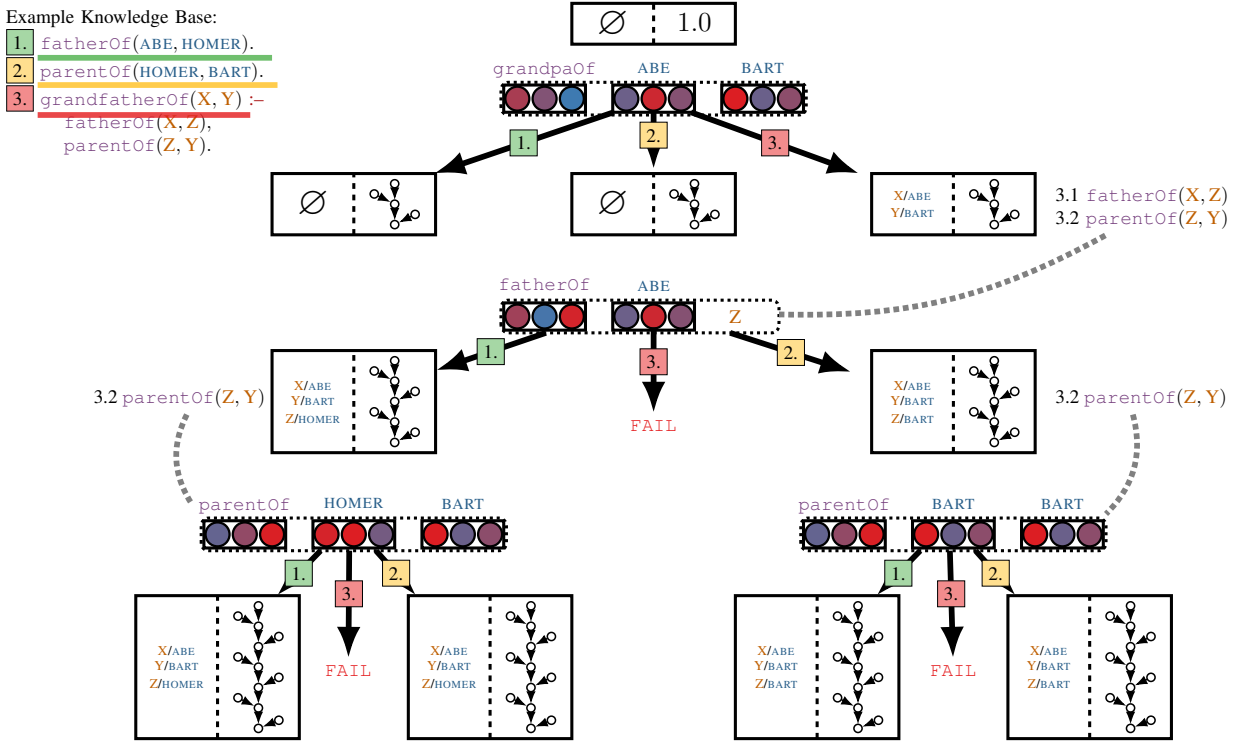
This continuous relaxation is at the crux of \glspl{NTP}' inability to scale to large datasets.
During both training and inference, \glspl{NTP} need to compute all possible proof trees needed for proving a query, relying on the continuous unification of the query with \emph{all} the rules and facts in the \gls{KB}.
This procedure quickly becomes infeasible for large datasets, as the number of nodes of the resulting computation graph grows exponentially.
Our insight is that we can radically reduce the computational complexity of inference and learning by generating only the most promising proof paths.
In particular, we show that the problem of finding the facts in the \gls{KB} that best explain a query can be reduced to a $k$-nearest neighbour problem, for which efficient exact and approximate solutions exist~\citep{DBLP:journals/corr/LiZSWZL16}.
This enables us to apply \glspl{NTP} to previously unreachable real-world datasets, such as WordNet.
\section{Background}
In \glspl{NTP}, the neural network structure is built recursively, and its construction is defined in terms of \emph{modules} similarly to dynamic neural module networks~\citep{DBLP:conf/naacl/AndreasRDK16}.
Each module, given a goal, a \gls{KB}, and a current proof state as inputs, produces a list of new proof states, where the proof states are neural networks representing partial proof success scores.
{\bf Unification Module.}
In backward chaining, unification between two atoms is used for checking whether they can represent the same structure.
In discrete unification, non-variable symbols are checked for equality, and the proof fails if the symbols differ.
In \glspl{NTP}, rather than comparing symbols, their \emph{embedding representations} are compared by means of a \gls{RBF} kernel.
This allows matching different symbols with similar semantics, such as matching relations like $\rel{grandFatherOf}$ and $\rel{grandpaOf}$.
Given a \emph{proof state} $\state = (\state_\subs, \state_\success)$, where $\state_\subs$ and $\state_\success$ denote a \emph{substitution set} and a \emph{proof score}, respectively, unification is computed as follows:
\begin{figure*}[h!]
    \centering
    \resizebox{0.9\textwidth}{!}{
      \includegraphics{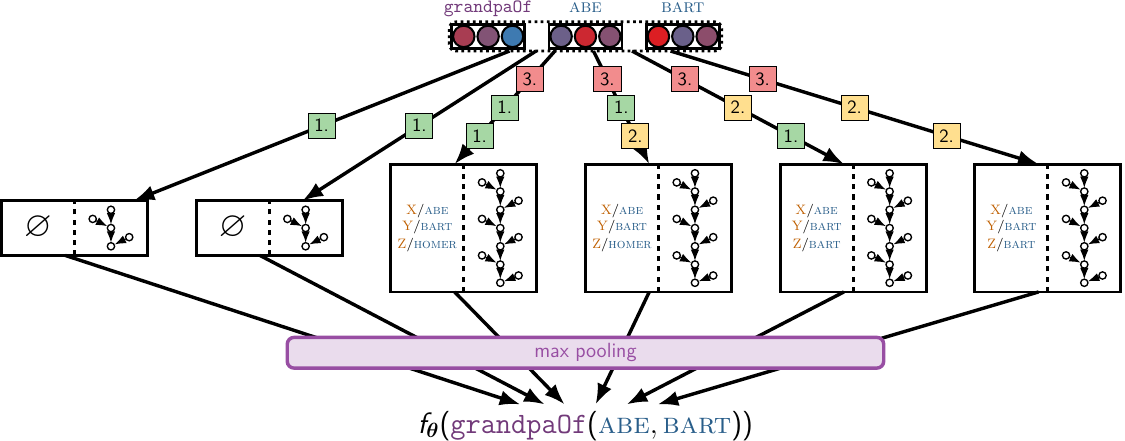}
    }
\caption{Depiction of the proof aggregation for the computation graph presented in \Cref{fig:overview}. %
Proof states resulting from the computation graph construction are all aggregated to obtain the final success score of proving a query.} \label{fig:aggregation}
\end{figure*}
\vspace{-0.5cm}
{\footnotesize
\begin{flalign*}
1. & \ \ \module{unify}_\params(\emptylist, \emptylist, \state) = \state&\\
2. & \ \ \module{unify}_\params(\emptylist, \lst{G}, \state) = \fail&\\
3. & \ \ \module{unify}_\params(\lst{H}, \emptylist, \state) = \fail&\\
4. & \ \ \module{unify}_\params(h::\lst{H}, g::\lst{G}, \state) = \module{unify}_\params(\lst{H},\lst{G},\state') &\\
& \ \ \ \ \text{with} \ \ \state' = (\state'_\subs, \state'_\success) \ \ \text{where:} &\\
\end{flalign*}
%
\vspace{-1.0cm}
%
\begin{align*}
\state'_\subs = & \state_\subs\union
\left\{\begin{array}{ll}
\{h/g\}         & \text{if } h\in \set{V}\\
\{g/h\}         & \text{if } g\in \set{V}, h\not\in \set{V}\\
\emptyset   & \text{otherwise}
\end{array}\right\} \\
\state'_\success = & \min\left(
\state_\success,
\left\{
    \begin{array}{ll}
        \krnl\left(\params_{h:}, \params_{g:}\right) & \text{if } h\not\in \set{V}, g\not\in \set{V}\\
        1 & \text{otherwise}
    \end{array}\right\}
\right) \\
\end{align*}
}
\vspace{-1.0cm}
\noindent where $\params_{h:}$ and $\params_{g:}$ denote the embedding representations of $h$ and $g$, respectively.
{\bf OR Module.}
This module attempts to apply rules in a \gls{KB}.
The name of this module stems from the fact that a \gls{KB} can be seen as a large disjunction of rules and facts.
In backward chaining reasoning systems, the OR module is used for unifying a goal with all facts and rules in a \gls{KB}: if the goal unifies with the head of the rule, then a series of goals is derived from the body of such a rule.
In \glspl{NTP}, we calculate the similarity between the rule and the facts via the \module{unify} operator.
Upon calculating the continuous unification scores, OR calls AND to prove all sub-goals in the body of the rule.

\vspace{-0.5cm}
{\footnotesize
\begin{flalign*}
\module{or}^{\kb}_\params(\lst{G}, d, \state) = \xs{\state' \ |\ & \state' \in \module{and}^{\kb}_\params(\lst{B}, d, \module{unify}_\params(\lst{H}, \lst{G}, \state)), \\
& \lst{H} \lif \lst{B} \in \kb}
\end{flalign*}
}
\vspace{-0.5cm}

{\bf AND Module.}
This module is used for proving a conjunction of sub-goals derived from a rule body.
It first applies substitutions to the first atom, which is afterwards proven by calling the OR module.
Remaining sub-goals are proven by recursively calling the AND module.

\vspace{-0.5cm}
{\footnotesize
\begin{flalign*}
1. &\ \ \module{and}^{\kb}_\params(\_, \_, \fail) = \fail&\\
2. &\ \ \module{and}^{\kb}_\params(\_, 0, \_) = \fail&\\
3. &\ \ \module{and}^{\kb}_\params(\emptylist, \_, \state) = \state&\\
4. &\ \ \module{and}^{\kb}_\params(\ls{G}:\lss{G}, d, \state) = \xs{\state''\ |\ \state''\in\module{and}^{\kb}_\params(\lss{G}, d, \state'), \\ 
& \ \ \ \ \state' \in \module{or}^{\kb}_\params(\fun{substitute}(\ls{G}, \state_\subs), d-1, \state)}&
\end{flalign*}
}
\vspace{-0.5cm}

For further details on NTPs and the particular implementation of these modules, see~\citet{rocktaschel2017end}

After building all the proof states, NTPs define the final success score of proving a query as an $\argmax$ over all the generated valid proof scores (neural networks).

\begin{example} \label{ex:inference}
Assume a \glspl{KB} $\mathcal{K}$, composed of $|\mathcal{K}|$ facts and no rules, for brevity.
Note that $|\mathcal{K}|$ can be impractical within the scope of \gls{NTP}.
For instance, Freebase~\citep{DBLP:conf/sigmod/BollackerEPST08} is composed of approximately 637 million facts, while YAGO3~\citep{DBLP:conf/cidr/MahdisoltaniBS15} is composed by approximately 9 million facts.
Given a query $g \triangleq [\rel{grandpaOf}, \const{abe}, \const{bart}]$, NTP compares its embedding representation  -- given by the embedding vectors of $\rel{grandpaOf}$, $\const{abe}$, and $\const{bart}$ -- with the representation of each of the $|\mathcal{K}|$ facts.
The resulting proof score of $g$ is given by:
\begin{equation} \label{eq:inference}
\begin{aligned}
 \max_{f \in \mathcal{K}} & \; \module{unify}_\params(g, [f_{p}, f_{s}, f_{o}], (\emptyset, \success)) \\ 
 & = \max_{f \in \mathcal{K}} \; \min \big\{
    \success,
    \krnl(\params_{\scriptsize\rel{grandpaOf}:}, \params_{f_{p}:}),\\
 &\qquad\qquad\qquad
    \krnl(\params_{\const{abe}:}, \params_{f_{s}:}),
    \krnl(\params_{\const{bart}:}, \params_{f_{o}:})
 \big\},
\end{aligned}
\end{equation}
\noindent where $f \triangleq [f_{p}, f_{s}, f_{o}]$ is a fact in $\mathcal{K}$ denoting a relationship of type $f_{p}$ between $f_{s}$ and $f_{o}$, $\params_{s:}$ is the embedding representation of a symbol $s$, $\success$ denotes the initial proof score, and $\krnl({}\cdot{}, {}\cdot{})$ denotes the \gls{RBF} kernel.
Note that the maximum proof score is given by the fact $f \in \mathcal{K}$ that maximises the similarity between its components and the goal $g$: solving the maximisation problem in \Cref{eq:inference} can be equivalently stated as a \emph{nearest neighbour search} problem.
In this work, we use \gls{ANNS} during the forward pass for considering only the most promising proof paths during the construction of the neural network.
\end{example}
\section{Nearest Neighbourhood Search}
From \Cref{ex:inference}, we can see that the inference problem can be reduced to a nearest neighbour search problem.
Given a query $g$, the problem is finding the fact(s) in $\mathcal{K}$ that maximise the unification score.
This represents a computational bottleneck, since it is very costly to find the exact nearest neighbour in high-dimensional Euclidean spaces, due to the curse of dimensionality~\citep{DBLP:conf/stoc/IndykM98}.
Exact methods are rarely more efficient than brute-force linear scan methods when the dimensionality is high~\citep{DBLP:journals/pami/GeHK014,DBLP:journals/corr/MalkovY16}.
A practical solution consists in \gls{ANNS} algorithms, which relax the condition of the exact search by allowing a small number of mistakes.
Several families of \gls{ANNS} algorithms exist, such as \gls{LSH}~\citep{DBLP:conf/nips/AndoniILRS15}, \gls{PQ}~\citep{DBLP:journals/pami/JegouDS11}, and \glspl{PG}~\citep{DBLP:journals/is/MalkovPLK14}.
In this work we use \gls{HNSW}~\citep{DBLP:journals/corr/MalkovY16,DBLP:conf/sisap/BoytsovN13}, a graph-based incremental \gls{ANNS} structure which can offer much better logarithmic complexity scaling in comparison with other approaches.
\section{Related Work} \label{sec:related}

Many machine learning methods rely on efficient nearest neighbour search for solving specific sub-problems.
Given the computational complexity of nearest neighbour search, approximate methods, driven by advanced index structures, hash or even graph-based approaches are used to speed up the bottleneck of costly comparison.
\gls{ANNS} algorithms have been used to speed up various sorts of machine learning models, including mixture model clustering~\citep{moore1999very}, case-based reasoning~\citep{wess1993using} to Gaussian process regression~\citep{shen2006fast}, among others.
Similarly to this work, \citet{rae2016scaling} also rely on approximate nearest neighbours to speed up Memory-Augmented neural networks.
Similarly to our work, they apply \gls{ANNS} to query the external memory (in our case the KB memory) for $k$ closest words.
They present drastic savings in speed and memory usage.
Though as of this moment, our speed savings are not as drastic, the memory savings we achieve are sufficient so that we can train on WordNet, a dataset previously considered out of reach of \glspl{NTP}.
\begin{table}[t]
    \centering
    \caption{AUC-PR results on Countries and MRR and HITS@$m$ on Kinship, Nations, and UMLS.}
    \label{tab:results}    
    \resizebox{\columnwidth}{!}{        
    \begin{tabular}{lrlcc}
        \toprule
        \multicolumn{2}{c}{\bf Dataset} & \multicolumn{1}{c}{\bf Metric} & \multicolumn{2}{c}{\bf Model} \\
        & & & \multicolumn{1}{c}{\textbf{NTP}} & \multicolumn{1}{c}{\textbf{NTP 2.0} ($k$ = 1)} \\
        \midrule
        \multirow{3}{*}{\bf Countries}
        & S1 & AUC-PR & $90.83 \pm 15.4$  & $\bm{97.04} \pm 4.47$ \\
        & S2 & AUC-PR & $87.40 \pm 11.7$  & $\bm{90.92} \pm 4.44$ \\
        & S3 & AUC-PR & $56.68 \pm 17.6$  & $\bm{85.55} \pm 7.10$ \\
        \midrule
        \multirow{3}{*}{\bf Kinship}
        && MRR & $0.60$& $\bm{0.65}$\\
        && HITS@1 & $0.48$ & $\bm{0.57}$\\
        && HITS@3 & $\bm{0.70}$ & $0.69$\\
        && HITS@10 & $0.78$ & $\bm{0.81}$\\
        \midrule
        \multirow{3}{*}{\bf Nations}
        && MRR & $0.75$ & $\bm{0.81}$\\
        && HITS@1 & $0.62$ & $\bm{0.73}$\\
        && HITS@3 & $\bm{0.86}$ & $0.83$\\
        && HITS@10 & $\bm{0.99}$ & $\bm{0.99}$\\
        \midrule
        \multirow{3}{*}{\bf UMLS}
        && MRR & $\bm{0.88}$ & $0.76$\\
        && HITS@1 & $\bm{0.82}$ & $0.68$\\
        && HITS@3 & $\bm{0.92}$ & $0.81$\\
        && HITS@10 & $\bm{0.97}$ & $0.88$\\

        \bottomrule
    \end{tabular}
}
\vspace{-0.5cm}
\end{table}
\begin{table}[t]
    \centering
    \caption{Rules induced on WordNet, with a confidence above $0.5$.}
    \label{tab:rules}
    \resizebox{\columnwidth}{!}{%
    \begin{tabular}{cl}
    \toprule
    \multicolumn{1}{c}{\bf Confidence} & \multicolumn{1}{l}{\bf Rule}\\
    \midrule
    %
    %
    0.584      & \rel{\_domain\_topic}(\var{X}, \var{Y}) \lif \rel{\_domain\_topic}(\var{Y}, \var{X}) \\
    0.786      & \rel{\_part\_of}(\var{X}, \var{Y}) \lif \rel{\_domain\_region}(\var{Y}, \var{X}) \\
    0.929      & \rel{\_similar\_to}(\var{X}, \var{Y}) \lif \rel{\_domain\_topic}(\var{Y}, \var{X}) \\
    0.943      & \rel{\_synset\_domain\_topic}(\var{X}, \var{Y}) \lif \rel{\_domain\_topic}(\var{Y}, \var{X}) \\
    0.998      & \rel{\_has\_part}(\var{X}, \var{Y}) \lif \rel{\_similar\_to}(\var{Y}, \var{X}) \\
    0.995      & \rel{\_member\_meronym}(\var{X}, \var{Y}) \lif \rel{\_member\_holonym}(\var{Y}, \var{X}) \\
    0.904      & \rel{\_domain\_topic}(\var{X}, \var{Y}) \lif \rel{\_has\_part}(\var{Y}, \var{X}) \\
    0.814      & \rel{\_member\_meronym}(\var{X}, \var{Y}) \lif \rel{\_member\_holonym}(\var{Y}, \var{X}) \\
    0.888      & \rel{\_part\_of}(\var{X}, \var{Y}) \lif \rel{\_domain\_topic}(\var{Y}, \var{X}) \\
    0.996      & \rel{\_member\_holonym}(\var{X}, \var{Y}) \lif \rel{\_member\_meronym}(\var{Y}, \var{X}) \\
    0.877      & \rel{\_part\_of}(\var{X}, \var{Y}) \lif \rel{\_domain\_topic}(\var{Y}, \var{X}) \\
    0.945      & \rel{\_synset\_domain\_topic}(\var{X}, \var{Y}) \lif \rel{\_domain\_region}(\var{Y}, \var{X}) \\
    0.879      & \rel{\_part\_of}(\var{X}, \var{Y}) \lif \rel{\_domain\_topic}(\var{Y}, \var{X}) \\
    0.926      & \rel{\_domain\_topic}(\var{X}, \var{Y}) \lif \rel{\_domain\_topic}(\var{Y}, \var{X}) \\
    0.995      & \rel{\_has\_instance}(\var{X}, \var{Y}) \lif \rel{\_type\_of}(\var{Y}, \var{X}) \\
    0.996      & \rel{\_type\_of}(\var{X}, \var{Y}) \lif \rel{\_has\_instance}(\var{Y}, \var{X}) \\

    \bottomrule
    \end{tabular}%
    }
\vspace{-0.5cm}
\end{table}
\section{Experiments} \label{sec:experiments}
We compared results obtained by our model, which we refer to as \gls{NTP} 2.0, with those obtained by the original \gls{NTP} proposed by \citet{rocktaschel2017end}.
Results on several smaller datasets -- namely Countries, Nations, Kinship, and UMLS -- are shown in \Cref{tab:results}.
When unifying goals with facts in the \gls{KB}, for each goal, we use \gls{ANNS} for retrieving the $k$ most similar (in embedding space) facts, and use those for computing the final proof scores.
We report results for $k = 1$, as we did not notice sensible differences for $k \in \{ 2, 5, 10 \}$.
However, we noticed sensible improvements in the case of Countries, and an overall decrease in performance in UMLS.
A possible explanation is that \gls{ANNS} (with $k = 1$), due to its inherently approximate nature, does not always retrieve the closest fact(s) exactly.
This behaviour may be a problem in some datasets where exact nearest neighbour search is crucial for correctly answering queries.
We also evaluated \gls{NTP} 2.0 on WordNet~\citep{DBLP:journals/cacm/Miller95}, a \gls{KB} encoding lexical knowledge about the English language. 
In particular, we use the WordNet used by \citet{DBLP:conf/nips/SocherCMN13} for their experiments.
This dataset is significantly larger than the other datasets used by \citet{rocktaschel2017end} -- it is composed by 38.696 entities, 11 relations, and the training set is composed by 112,581 facts.
In WordNet, the accuracies on the validation and test sets were 65.29\% and 65.72\%, respectively -- which is on par with the Distance Model, a Neural Link Predictor discussed by~\citet{DBLP:conf/nips/SocherCMN13}, which achieves a test accuracy of 68.3\%.
However, we did not consider a full hyper-parameter sweep, and did not regularise the model using Neural Link Predictors, which sensibly improves \glspl{NTP}' predictive accuracy~\citep{rocktaschel2017end}.
A subset of the induced rules is shown in \Cref{tab:rules}.
\section{Conclusions}
We proposed a way to sensibly scale up \glspl{NTP} by reducing parts of their inference steps to \gls{ANNS} problems, for which very efficient and scalable solutions exist in the literature.

\small
\bibliographystyle{unsrtnat}
\bibliography{scaling-ntp}

\begin{thebibliography}{35}
\providecommand{\natexlab}[1]{#1}
\providecommand{\url}[1]{\texttt{#1}}
\expandafter\ifx\csname urlstyle\endcsname\relax
  \providecommand{\doi}[1]{doi: #1}\else
  \providecommand{\doi}{doi: \begingroup \urlstyle{rm}\Url}\fi

\bibitem[Rockt\"{a}schel and Riedel(2017)]{rocktaschel2017end}
Tim Rockt\"{a}schel and Sebastian Riedel.
\newblock End-to-end differentiable proving.
\newblock In \emph{Advances in Neural Information Processing Systems 30}, pages
  3788--3800. 2017.

\bibitem[Shen(1988)]{shen1988theoretical}
ZL~Shen.
\newblock A theoretical framework of fuzzy prolog machine.
\newblock \emph{Fuzzy Computing-Theory, Hardware and Applications}, 1988.

\bibitem[Ding et~al.(1996)Ding, Teh, Wang, and Lui]{ding1996prolog}
Liya Ding, Hoon~Heng Teh, Peizhuang Wang, and Ho~Chung Lui.
\newblock A prolog-like inference system based on neural logic—an attempt
  towards fuzzy neural logic programming.
\newblock \emph{Fuzzy Sets and Systems}, 82\penalty0 (2):\penalty0 235--251,
  1996.

\bibitem[Garcez et~al.(2012)Garcez, Broda, and Gabbay]{garcez2012neural}
Artur S~d'Avila Garcez, Krysia~B Broda, and Dov~M Gabbay.
\newblock \emph{Neural-symbolic learning systems: foundations and
  applications}.
\newblock Springer Science \& Business Media, 2012.

\bibitem[Marcus(2018)]{DBLP:journals/corr/abs-1801-00631}
Gary Marcus.
\newblock Deep learning: {A} critical appraisal.
\newblock \emph{CoRR}, abs/1801.00631, 2018.

\bibitem[Raedt and Kersting(2008)]{DBLP:conf/ilp/RaedtK08}
Luc~De Raedt and Kristian Kersting.
\newblock Probabilistic inductive logic programming.
\newblock In \emph{Probabilistic Inductive Logic Programming - Theory and
  Applications}, volume 4911 of \emph{Lecture Notes in Artificial
  Intelligence}, pages 1--27. Springer, 2008.

\bibitem[Bengio et~al.(2013)Bengio, Courville, and
  Vincent]{DBLP:journals/pami/BengioCV13}
Yoshua Bengio, Aaron~C. Courville, and Pascal Vincent.
\newblock Representation learning: {A} review and new perspectives.
\newblock \emph{{IEEE} Trans. Pattern Anal. Mach. Intell.}, 35\penalty0
  (8):\penalty0 1798--1828, 2013.

\bibitem[Graves et~al.(2014)Graves, Wayne, and
  Danihelka]{DBLP:journals/corr/GravesWD14}
Alex Graves, Greg Wayne, and Ivo Danihelka.
\newblock Neural turing machines.
\newblock \emph{CoRR}, abs/1410.5401, 2014.

\bibitem[Graves et~al.(2016)Graves, Wayne, Reynolds, Harley, Danihelka,
  Grabska{-}Barwinska, Colmenarejo, Grefenstette, Ramalho, Agapiou, Badia,
  Hermann, Zwols, Ostrovski, Cain, King, Summerfield, Blunsom, Kavukcuoglu, and
  Hassabis]{DBLP:journals/nature/GravesWRHDGCGRA16}
Alex Graves, Greg Wayne, Malcolm Reynolds, Tim Harley, Ivo Danihelka, Agnieszka
  Grabska{-}Barwinska, Sergio~Gomez Colmenarejo, Edward Grefenstette, Tiago
  Ramalho, John Agapiou, Adri{\`{a}}~Puigdom{\`{e}}nech Badia, Karl~Moritz
  Hermann, Yori Zwols, Georg Ostrovski, Adam Cain, Helen King, Christopher
  Summerfield, Phil Blunsom, Koray Kavukcuoglu, and Demis Hassabis.
\newblock Hybrid computing using a neural network with dynamic external memory.
\newblock \emph{Nature}, 538\penalty0 (7626):\penalty0 471--476, 2016.

\bibitem[Kaiser and Sutskever(2016)]{DBLP:journals/corr/KaiserS15}
Lukasz Kaiser and Ilya Sutskever.
\newblock Neural gpus learn algorithms.
\newblock In \emph{Proceedings of the International Conference on Learning
  Representations}, 2016.

\bibitem[Neelakantan et~al.(2016)Neelakantan, Le, and
  Sutskever]{DBLP:journals/corr/NeelakantanLS15}
Arvind Neelakantan, Quoc~V. Le, and Ilya Sutskever.
\newblock Neural programmer: Inducing latent programs with gradient descent.
\newblock In \emph{Proceedings of the International Conference on Learning
  Representations}, 2016.

\bibitem[Andrychowicz and Kurach(2016)]{DBLP:journals/corr/AndrychowiczK16}
Marcin Andrychowicz and Karol Kurach.
\newblock Learning efficient algorithms with hierarchical attentive memory.
\newblock \emph{CoRR}, abs/1602.03218, 2016.

\bibitem[Reed and de~Freitas(2016)]{DBLP:journals/corr/ReedF15}
Scott~E. Reed and Nando de~Freitas.
\newblock Neural programmer-interpreters.
\newblock In \emph{Proceedings of the International Conference on Learning
  Representations (ICLR)}, 2016.

\bibitem[Bosnjak et~al.(2017)Bosnjak, Rockt{\"{a}}schel, Naradowsky, and
  Riedel]{DBLP:conf/icml/BosnjakRNR17}
Matko Bosnjak, Tim Rockt{\"{a}}schel, Jason Naradowsky, and Sebastian Riedel.
\newblock Programming with a differentiable forth interpreter.
\newblock In \emph{Proceedings of the 34th International Conference on Machine
  Learning, {ICML}}, volume~70, pages 547--556. {PMLR}, 2017.

\bibitem[Gaunt et~al.(2016)Gaunt, Brockschmidt, Singh, Kushman, Kohli, Taylor,
  and Tarlow]{gaunt2016terpret}
Alexander~L Gaunt, Marc Brockschmidt, Rishabh Singh, Nate Kushman, Pushmeet
  Kohli, Jonathan Taylor, and Daniel Tarlow.
\newblock Terpret: A probabilistic programming language for program induction.
\newblock \emph{arXiv preprint arXiv:1608.04428}, 2016.

\bibitem[Parisotto et~al.(2016)Parisotto, Mohamed, Singh, Li, Zhou, and
  Kohli]{parisotto2016neuro}
Emilio Parisotto, Abdel-rahman Mohamed, Rishabh Singh, Lihong Li, Dengyong
  Zhou, and Pushmeet Kohli.
\newblock Neuro-symbolic program synthesis.
\newblock \emph{arXiv preprint arXiv:1611.01855}, 2016.

\bibitem[Evans and Grefenstette(2018)]{evans2018learning}
Richard Evans and Edward Grefenstette.
\newblock Learning explanatory rules from noisy data.
\newblock \emph{Journal of Artificial Intelligence Research}, 61:\penalty0
  1--64, 2018.

\bibitem[Nickel et~al.(2016)Nickel, Murphy, Tresp, and
  Gabrilovich]{DBLP:journals/pieee/Nickel0TG16}
Maximilian Nickel, Kevin Murphy, Volker Tresp, and Evgeniy Gabrilovich.
\newblock A review of relational machine learning for knowledge graphs.
\newblock \emph{Proceedings of the {IEEE}}, 104\penalty0 (1):\penalty0 11--33,
  2016.

\bibitem[Li et~al.(2016)Li, Zhang, Sun, Wang, Zhang, and
  Lin]{DBLP:journals/corr/LiZSWZL16}
Wen Li, Ying Zhang, Yifang Sun, Wei Wang, Wenjie Zhang, and Xuemin Lin.
\newblock Approximate nearest neighbor search on high dimensional data -
  experiments, analyses, and improvement (v1.0).
\newblock \emph{CoRR}, abs/1610.02455, 2016.

\bibitem[Andreas et~al.(2016)Andreas, Rohrbach, Darrell, and
  Klein]{DBLP:conf/naacl/AndreasRDK16}
Jacob Andreas, Marcus Rohrbach, Trevor Darrell, and Dan Klein.
\newblock Learning to compose neural networks for question answering.
\newblock In Kevin Knight et~al., editors, \emph{{NAACL} {HLT} 2016, The 2016
  Conference of the North American Chapter of the Association for Computational
  Linguistics: Human Language Technologies}, pages 1545--1554. The Association
  for Computational Linguistics, 2016.

\bibitem[Bollacker et~al.(2008)Bollacker, Evans, Paritosh, Sturge, and
  Taylor]{DBLP:conf/sigmod/BollackerEPST08}
Kurt~D. Bollacker, Colin Evans, Praveen Paritosh, Tim Sturge, and Jamie Taylor.
\newblock Freebase: a collaboratively created graph database for structuring
  human knowledge.
\newblock In \emph{Proceedings of the {ACM} {SIGMOD} International Conference
  on Management of Data, {SIGMOD}}, pages 1247--1250. {ACM}, 2008.

\bibitem[Mahdisoltani et~al.(2015)Mahdisoltani, Biega, and
  Suchanek]{DBLP:conf/cidr/MahdisoltaniBS15}
Farzaneh Mahdisoltani, Joanna Biega, and Fabian~M. Suchanek.
\newblock {YAGO3:} {A} knowledge base from multilingual wikipedias.
\newblock In \emph{{CIDR} 2015, Seventh Biennial Conference on Innovative Data
  Systems Research}, 2015.

\bibitem[Indyk and Motwani(1998)]{DBLP:conf/stoc/IndykM98}
Piotr Indyk and Rajeev Motwani.
\newblock Approximate nearest neighbors: Towards removing the curse of
  dimensionality.
\newblock In Jeffrey~Scott Vitter, editor, \emph{Proceedings of the Thirtieth
  Annual {ACM} Symposium on the Theory of Computing}, pages 604--613. {ACM},
  1998.

\bibitem[Ge et~al.(2014)Ge, He, Ke, and Sun]{DBLP:journals/pami/GeHK014}
Tiezheng Ge, Kaiming He, Qifa Ke, and Jian Sun.
\newblock Optimized product quantization.
\newblock \emph{{IEEE} Transactions on Pattern Analysis and Machine
  Intelligence}, 36\penalty0 (4):\penalty0 744--755, 2014.

\bibitem[Malkov and Yashunin(2016)]{DBLP:journals/corr/MalkovY16}
Yury~A. Malkov and D.~A. Yashunin.
\newblock Efficient and robust approximate nearest neighbor search using
  hierarchical navigable small world graphs.
\newblock \emph{CoRR}, abs/1603.09320, 2016.

\bibitem[Andoni et~al.(2015)Andoni, Indyk, Laarhoven, Razenshteyn, and
  Schmidt]{DBLP:conf/nips/AndoniILRS15}
Alexandr Andoni, Piotr Indyk, Thijs Laarhoven, Ilya~P. Razenshteyn, and Ludwig
  Schmidt.
\newblock Practical and optimal {LSH} for angular distance.
\newblock In Corinna Cortes et~al., editors, \emph{Advances in Neural
  Information Processing Systems 28: Annual Conference on Neural Information
  Processing Systems}, pages 1225--1233, 2015.

\bibitem[J{\'{e}}gou et~al.(2011)J{\'{e}}gou, Douze, and
  Schmid]{DBLP:journals/pami/JegouDS11}
Herv{\'{e}} J{\'{e}}gou, Matthijs Douze, and Cordelia Schmid.
\newblock Product quantization for nearest neighbor search.
\newblock \emph{{IEEE} Trans. Pattern Anal. Mach. Intell.}, 33\penalty0
  (1):\penalty0 117--128, 2011.

\bibitem[Malkov et~al.(2014)Malkov, Ponomarenko, Logvinov, and
  Krylov]{DBLP:journals/is/MalkovPLK14}
Yury Malkov, Alexander Ponomarenko, Andrey Logvinov, and Vladimir Krylov.
\newblock Approximate nearest neighbor algorithm based on navigable small world
  graphs.
\newblock \emph{Inf. Syst.}, 45:\penalty0 61--68, 2014.

\bibitem[Boytsov and Naidan(2013)]{DBLP:conf/sisap/BoytsovN13}
Leonid Boytsov and Bilegsaikhan Naidan.
\newblock Engineering efficient and effective non-metric space library.
\newblock In \emph{Similarity Search and Applications - 6th International
  Conference, {SISAP} 2013, Proceedings}, pages 280--293, 2013.

\bibitem[Moore(1999)]{moore1999very}
Andrew~W Moore.
\newblock Very fast em-based mixture model clustering using multiresolution
  kd-trees.
\newblock In \emph{Advances in Neural information processing systems}, pages
  543--549, 1999.

\bibitem[Wess et~al.(1993)Wess, Althoff, and Derwand]{wess1993using}
Stefan Wess, Klaus-Dieter Althoff, and Guido Derwand.
\newblock Using kd trees to improve the retrieval step in case-based reasoning.
\newblock In \emph{European Workshop on Case-Based Reasoning}, pages 167--181.
  Springer, 1993.

\bibitem[Shen et~al.(2006)Shen, Seeger, and Ng]{shen2006fast}
Yirong Shen, Matthias Seeger, and Andrew~Y Ng.
\newblock Fast gaussian process regression using kd-trees.
\newblock In \emph{Advances in neural information processing systems}, pages
  1225--1232, 2006.

\bibitem[Rae et~al.(2016)Rae, Hunt, Danihelka, Harley, Senior, Wayne, Graves,
  and Lillicrap]{rae2016scaling}
Jack Rae, Jonathan~J Hunt, Ivo Danihelka, Timothy Harley, Andrew~W Senior,
  Gregory Wayne, Alex Graves, and Tim Lillicrap.
\newblock Scaling memory-augmented neural networks with sparse reads and
  writes.
\newblock In \emph{Advances in Neural Information Processing Systems}, pages
  3621--3629, 2016.

\bibitem[Miller(1995)]{DBLP:journals/cacm/Miller95}
George~A. Miller.
\newblock Wordnet: {A} lexical database for english.
\newblock \emph{Commun. {ACM}}, 38\penalty0 (11):\penalty0 39--41, 1995.

\bibitem[Socher et~al.(2013)Socher, Chen, Manning, and
  Ng]{DBLP:conf/nips/SocherCMN13}
Richard Socher, Danqi Chen, Christopher~D. Manning, and Andrew~Y. Ng.
\newblock Reasoning with neural tensor networks for knowledge base completion.
\newblock In Christopher J.~C. Burges et~al., editors, \emph{Advances in Neural
  Information Processing Systems 26: 27th Annual Conference on Neural
  Information Processing Systems}, pages 926--934, 2013.

\end{thebibliography}

\end{document}